\documentclass[10pt,twocolumn,letterpaper]{article}

\usepackage[pagenumbers]{wacv}      

\usepackage{graphicx}
\usepackage{amsmath}
\usepackage{amssymb}
\usepackage{booktabs}

\usepackage[pagebackref,breaklinks,colorlinks]{hyperref}

\usepackage[capitalize]{cleveref}
\crefname{section}{Sec.}{Secs.}
\Crefname{section}{Section}{Sections}
\Crefname{table}{Table}{Tables}
\crefname{table}{Tab.}{Tabs.}

\usepackage{xcolor}

\begin{document}

\title{MHB: Multimodal Handshape-aware Boundary Detection for Continuous Sign Language Recognition}

\author{
Mingyu Zhao$^{1}$,
Zhanfu Yang$^{1}$,
Yang Zhou$^{1}$,
Zhaoyang Xia$^{1}$,
Can Jin$^{1}$,
Xiaoxiao He$^{1}$,\\
Dimitris N.~Metaxas$^{1}$\\[2pt]
$^{1}$Rutgers University\\
$^{1}$110 Frelinghuysen Road, Piscataway, NJ 08854, USA\\
{\tt\small \{zhao.mingyu, zhanfu.yang, eta.yang, zx149, can.jin, xiaoxiao.he\}@rutgers.edu}\\
{\tt\small dnm@cs.rutgers.edu}\\
}
\maketitle

\begin{abstract}
This paper employs a multimodal approach for continuous sign recognition by first using ML for detecting the start and end frames of signs in videos of American Sign Language (ASL) sentences, and then by recognizing the segmented signs. For improved robustness we use 3D skeletal features extracted from sign language videos to take into account the convergence of sign properties and their dynamics that tend to cluster at sign boundaries. Another focus of this paper is the incorporation of information from 3D handshape for boundary detection. To detect handshapes normally expected at the beginning and end of signs, we pretrain a handshape classifier for detection of 87 linguistically defined canonical handshape categories using a dataset that we created by integrating and normalizing several existing datasets. A multimodal fusion module is then used to unify the pretrained sign video segmentation framework and handshape classification models. Finally, the estimated boundaries are used for sign recognition, where the recognition model is trained on a large database containing both citation-form isolated signs and signs pre-segmented (based on manual annotations) from continuous signing—as such signs often differ a bit in certain respects. We evaluate our method on the ASLLRP corpus and demonstrate significant improvements over previous work.
\end{abstract}


\vspace{-14pt}
\section{Introduction}
\label{sec:intro}
\vspace{-2pt}

Signed languages are fully developed visual-gestural languages used by Deaf and hard-of-hearing communities worldwide.  Unlike spoken languages, they convey meaning through hand shapes, orientations, movements, facial expressions, head gestures, and body posture \cite{brentari-phon}. American Sign Language (ASL) is the primary language used by more than 500,000 people in the United States \cite{howmany}. It is also ranked as the third most studied non-native language in the country \cite{mla-asl}. Despite advances in automatic speech and natural language processing, Sign Language Recognition (SLR) remains a complex problem for 2 main reasons: (1) the high-dimensional spatio-temporal dynamics of gestures, and (2) a general lack of large-scale annotated datasets \cite{huang2018video}. Most research to date has focused on Isolated Sign Language Recognition (ISLR), where each input clip contains exactly one sign. However, in real-world communication, signed languages are produced as continuous streams of signing, not isolated signs.

Compared with ISLR, Continuous Sign Language Recognition (CSLR) is more complex, in part because of significant differences in sign production in continuous signing as compared with citation forms \cite{neidle2023challenges}. Among many complicating factors in continuous signing are the potential independence of signing on the 2 hands, and the tendency for sign production to be influenced by the properties of adjacent segments. Note that in the current research, we are only attending to signs that include production by the dominant hand, i.e., 1-handed signs produced on the dominant hand and 2-handed signs.  We are not considering 1-handed signs produced on the non-dominant hand.

A major challenge in CSLR lies in the difficulty of locating precise temporal boundaries for each sign segment, i.e., identifying the start and end frames corresponding to each sign \cite{neidle2023challenges}.  Another major difficulty arises from the fact that natural signing includes things that CANNOT just be looked up in a “dictionary” of isolated signs (even if they could be perfectly segmented)—including potentially: false starts, things like classifiers and gestures, which do not belong to any finite classes of elements, as well as fingerspelled signs, potentially corresponding to any word or proper name from the spoken language (often with letters missing or transposed). This is why any ultimate strategy for recognition of signs from continuous signing needs to be able to distinguish among types of signs, as different recognition strategies need to be applied to different types of signs \cite{yanovich2016detection}.  Note, however, that the research reported here does not do this, so we have submitted our segmented signs for sign recognition only in cases where we have at least 6 examples of those signs in our training dataset. Thus, we are not currently carrying out end-to-end sign recognition, although that is the long-term goal.  The sign recognition component of this research is intended to demonstrate the usefulness of segmentation for sign recognition.

Some prior work relies on relatively small datasets for experimentation, such
as \cite{gonzalez2011sign, moryossef2023linguistically, zhang2023handshape}. Other work (e.g. \cite{rastgoo2024transformer}) attempts to synthesize pseudo-continuous sign language datasets by concatenating isolated sign videos. However, none of these strategies has been applied to real large-scale continuous sign language datasets for boundary detection. Most approaches to sign boundary detection follow a similar 2-stage pipeline. They typically first extract visual features from raw videos using tools such as OpenPose or 3D convolutional neural networks, and then feed these features into deep learning models such as LSTMs or Transformers for temporal modeling and boundary prediction \cite{rastgoo2024transformer, renz2021sign, moryossef2023linguistically}.  These methods have shown some success, but they generally neglect 2 important perspectives: (1) multimodal insights from patterns involving handshape, motion, velocity, acceleration, facial expressions, upper body position, and head gestures, known to correlate with sign transitions, which have the potential to provide strong signals for boundary detection; and (2) the use of more specialized skeleton-based approaches for pose modeling, as most  methods rely on general-purpose deep models and overlook structured methods that are better suited to skeletal data.

Building on these observations, we propose a multimodal boundary detection framework, incorporating a pretrained handshape classification model with a sign boundary detection model. We use the estimated boundaries for sign recognition in a well-trained sign recognition model. We propose this framework and, as ongoing work, are extending the experimental coverage by incorporating additional datasets. This manuscript is under active revision.

There is no standard written representation for signs.  Researchers often use English-based ID glossing to identify signs, as we also do.  But a major challenge is ensuring consistency in gloss labeling, i.e., a 1-1 correspondence between signs and glosses, and many datasets fail to meet this requirement (e.g., \cite{neidle2022resources}). One major advantage of the datasets we are using is that we ensure consistency in gloss representations across all the datasets we use.

Our contributions are summarized as follows:
\vspace{-6pt}
\begin{itemize}

    \item We design a segmentation module based on a spatiotemporal convolutional network, incorporating velocity and acceleration features. This outperforms state-of-the-art methods in CSL boundary detection.
    \vspace{-7pt}
    \item We propose a novel method to enhance boundary detection by incorporating handshape information. A multimodal fusion module is introduced, which integrates features from a pretrained handshape classifier into the segmentation stream using a cross-attention mechanism with a gating mechanism.
    \vspace{-7pt}
    \item We present a new CSL recognition pipeline combining a state-of-the-art segmentation model with a sign language recognition model. Experiments demonstrate the effectiveness of our segmentation framework.
\end{itemize}

\vspace{-3pt}
\section{Related Work}
\label{sec:intro}
\vspace{-1pt}
\subsection{Isolated Sign Language Recognition}\label{subsec:problem_formulation}
\vspace{-1pt}
Recent advances in Isolated Sign Language Recognition (ISLR) have leveraged both visual and skeletal modalities to recognize signs from trimmed video clips. Early methods mostly relied on handcrafted features and Hidden Markov Models (HMMs) \cite{starner1998real}. With the rise of deep learning, convolutional neural networks (CNNs) \cite{pigou2015sign} and hierarchical attention networks (HANs) \cite{huang2018video} became widely used to model spatial and temporal dependencies in isolated signs.

More recent approaches adopt BiLSTM \cite{camgoz2017subunets}, Transformer-based architectures \cite{rastgoo2024transformer}, and graph-based models (e.g., ST-GCN \cite{zhou2024multimodal}), to capture spatiotemporal dynamics. Some work also explores multimodal fusion \cite{zhou2020spatial}, combining RGB, depth, optical flow, and skeletal data to improve recognition accuracy. In particular, skeletal features extracted from pose estimation models offer a compact and robust representation of sign gestures, leading to improved performance in data-limited scenarios \cite{li2020word}.

Despite impressive progress, ISLR methods typically assume well-segmented input clips, limiting  applicability to real-world CSL scenarios. This motivates the development of methods that can extend to continuous sign recognition.

\subsection{Continuous Sign Language Recognition}

Continuous Sign Language Recognition (CSLR) aims to segment and recognize signs from natural sign language sentences. This has important potential applications to real-world scenarios,  as it could provide useful input to facilitate such things as live interpretation or video captioning.

Early approaches relied more on alignment-based strategies, which offered a good way to model temporal sequences. Vogler and Metaxas \cite{vogler1999parallel,vogler2001framework} proposed parallel HMM frameworks that modeled the simultaneous movements of multiple articulators (e.g., hands and facial expressions), enabling recognition of continuous signing without requiring manual segmentation. Such work laid the foundation for structured temporal modeling in CSLR. Deep learning methods introduced alignment-free techniques such as Connectionist Temporal Classification (CTC) \cite{graves2006connectionist}, which allowed sequence-level supervision without the need for frame-level annotations.

Later research introduced iterative alignment schemes using RNNs or TCNs to refine temporal segmentation. More recently, to address the intrinsic ambiguity in gloss-to-frame alignment, several studies propose incorporating linguistic priors and contextual constraints.  For example, Zhang et al. \cite{zhang2023c2st} introduce a gloss-conditioned decoder  integrating prior gloss context into the decoding process. Guo et al. \cite{guo2023distilling} enhance temporal modeling via cross-temporal knowledge distillation. Others also introduce gloss-level embeddings as guidance for visual feature learning \cite{guo2024gloss}.

\subsection{Sign Boundary Detection}\label{subsec:problem_formulation}

For CSLR, it is essential to identify the precise temporal boundaries of individual signs within continuous video streams. Early approaches treated boundary detection as a frame-wise classification problem or relied on dynamic programming to align gloss-level annotations with video frames \cite{gonzalez2011sign}. Still, these methods overlook the holistic structure of signed sentences. Subsequent methods leveraged temporal convolutional networks such as (TCNs) \cite{renz2021sign} to better capture long-range dependencies in signing sequences. More recently, Transformer-based architectures have been proposed. For example, Moryossef \cite{moryossef2023linguistically} encodes pose feature with an LSTM and applies a greedy segmentation decoder to identify sign units. Rastgoo \cite{rastgoo2024transformer} proposes a Transformer-based framework, but it relies on artificially constructed rather than real CSL data. Zhang \cite{zhang2023handshape} designs dual encoders for glosses and handshapes, fusing those representations via a joint head for gloss prediction. Recently, Rastgoo \cite{rastgoo2025non} employs a GCN-based model to extract skeletal features that are then embedded into a Transformer-based encoder for boundary detection. These approaches demonstrate strong performance in modeling long-range dependencies, but still overlook the integration of richer multimodal information, such as motion patterns and acceleration, which could provide complementary signals for more accurate boundary detection and recognition.


\section{Method}
\label{sec:intro}

\subsection{Problem Formulation}

In this work, we formulate the sign language segmentation task as a frame-wise BIO annotation problem over continuous sign language video sequences. Given a sequence of video frames $X = \{x_1, x_2, ..., x_N\}$, the goal is to predict a corresponding label sequence $L = \{l_1, l_2, ..., l_N\} \in \{B, I, O\}^N $, where the label "B" indicating the boundary frame of a sign segment, "I" indicating that the frame lies within a sign, specifically between the start frame and the end frame, and "O" indicating a frame that is outside any sign segment. This BIO annotation is better suited for modeling the continuous sign language segmentation task than traditional IO tagging, as sign segments in continuous sign language video often follow one another without any O frames in between, which results in these two adjacent segments being merged and counted as one segment\cite{moryossef2023linguistically}.

\subsection{Spatio-Temporal Convolutional Segmentation}

\begin{figure*}[ht]
\centering
\includegraphics[width=\linewidth]{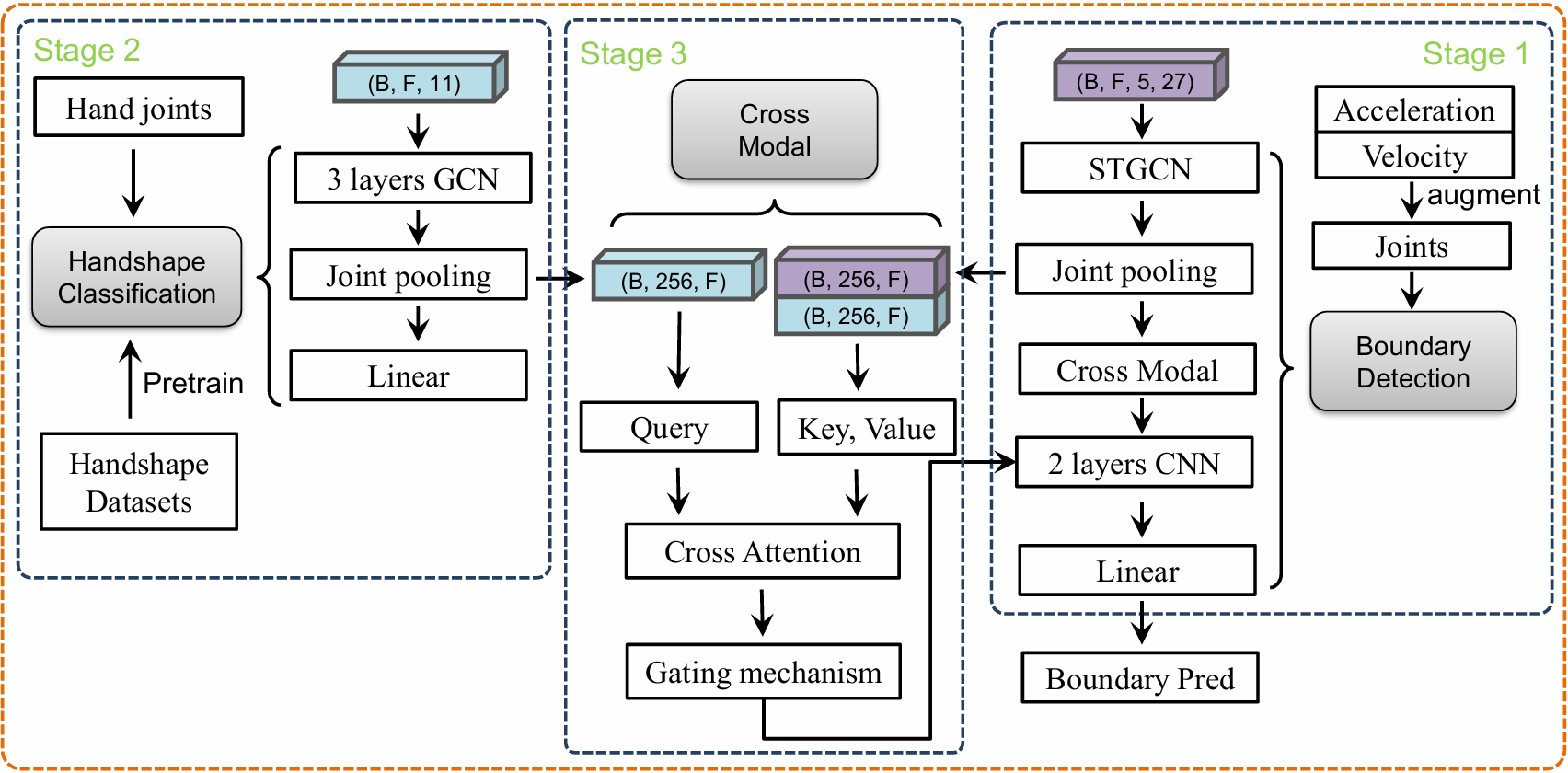}
\caption{Overview of our proposed multimodal boundary detection. The left branch shows the handshape classification module, which is pretrained on curated handshape datasets using a 3-layer GCN and later used to extract frame-wise handshape features. We pretrain this branch in stage 2. The right branch is the segmentation network based on ST-GCN and CNN modules, which processes all skeletal joints augmented with their velocity and acceleration. We pretrain this branch without the Cross-Modal layer in stage 1. In the middle, a cross-modal attention module with a gating mechanism fuses the handshape and segmentation features to enhance temporal boundary prediction. We fine-tune the right branch with the cross-modal attention module in stage 3. The final segmentation output is further used to extract sign clips and then used for sign recognition.
}
\label{fig:stgcn}
\end{figure*} 

We define S(x) as the skeleton extracted model\cite{fang2022alphapose}. From a video frame sequence $X = \{x_1, x_2, ..., x_N\}$, we extract the corresponding frame-wise 27 skeleton joint representations $Y = \{Y_1, Y_2, ..., Y_N\}$, where $Y_i = S(x_i)$. To meet input requirements of the STGCN-based model, we pad all extracted skeleton sequences along the temporal dimension to a fixed length of \text{T} frames. Ground-truth labels are extended to include 4 classes: \text{B}, \text{I}, \text{O}, and \text{P} (padding), corresponding to label indices 3, 2, 1, and 0, respectively. 

The extracted skeleton sequences $Y = \{Y_1, Y_2, ..., Y_T\}$ are then fed into a Spatio-Temporal Convolutional Segmentation Network. As shown in Figure~\ref{fig:stgcn}, the front part of the network is constructed based on the ST-GCN framework proposed by Yan \etal \cite{han2020stgcn}, which effectively models the spatio-temporal dynamics of human motion. Specifically, we stack 10 ST-GCN layers to progressively capture high-level spatio-temporal patterns across the skeleton sequence.

The output features from the final ST-GCN layer are then passed into a segmentation module designed for frame-wise prediction. To aggregate information across joints, we first apply a spatial pooling operation over the 27 joints at each time step. The pooled features are then processed through 2 temporal convolutional layers, followed by a linear projection layer that produces frame-level predictions over the segmentation labels. This design allows the model to first learn rich spatial and temporal dependencies using the ST-GCN backbone and then map the encoded motion features to fine-grained frame-wise segmentation outputs.

We adopt a combined loss function consisting of a frame-wise prediction loss and a boundary loss. 

The frame-wise prediction loss is defined as a weighted cross-entropy loss. Specifically, we assign a higher class weight to boundary frames (class 3) and a lower weight to background frames (class 0), in order to emphasize the importance of accurate boundary detection and reduce the influence of dominant background frames. The loss is computed as:
\vspace{-12pt}
\[
\mathcal{L}_{\text{frame}} = - \sum_{t=1}^{T} w_{y_t} \cdot \log \left( \hat{y}_t[y_t] \right),
\]
\noindent
where \( \hat{y}_t \in \mathbb{R}^C \) is the predicted class probability at time \( t \), \( y_t \in \{0, 1, 2, 3\} \) is the ground-truth label, and \( w_{y_t} \) is the class-specific weight.

The \text{boundary loss} is defined as the absolute difference between the number of predicted boundaries and ground-truth boundaries, averaged over the utterance:

\[
\mathcal{L}_{\text{boundary}} = \frac{1}{U} \sum_{u=1}^{U} \left| N_{\text{pred}}^{(u)} - N_{\text{true}}^{(u)} \right|,
\]
\noindent
where \( N_{\text{pred}}^{(u)} \) and \( N_{\text{true}}^{(u)} \) represent the number of predicted and true boundaries in the \( u \)-th utterance, and \( U \) is the number of all utterances.

The final loss $L_{seg}$ is a weighted combination of the 2 components:
\vspace{-3pt}
\[
\mathcal{L}_{\text{seg}} = \mathcal{L}_{\text{frame}} + \lambda \cdot \mathcal{L}_{\text{boundary}},
\]
\noindent
where \( \lambda \) is a hyperparameter controlling the contribution of the boundary loss.

\subsection{Handshape Classification}
\vspace{-3pt}
We formulate the handshape recognition task as a frame-wise classification problem over hand regions in sign video sequences. Given a video frame sequence
$X = \{x_1, x_2, ..., x_N\}$, we extract the corresponding frame-wise hand joint representations $X_H = \{x^h_1, x^h_2, ..., x^h_N\}$. We use the established set of 87 canonical handshape categories, indexed from 0 to 86. Rather than assigning an absolute label to each frame, the goal is to estimate the degree of similarity between the observed handshape and these canonical categories. Thus, we want to find such a handshape classifier denoted as $H(x_H) \in \{0,1,...,86\}^N$. However, in practice, some frames exhibit blurred or occluded hand configurations. To address this, we manually sample a subset of clear frames to construct a high-quality training dataset denoted as $X^S_H = \{x_{h^s_1}, x_{h^s_2}, ..., x_{h^s_k}\}$, where $h^s_1, h^s_2,...,h^s_k$ are selected k frames. The handshape classifier $H(x^H)$ is then pretrained on this dataset and integrated into our multimodal segmentation framework. See Section~\ref{sec:handshape_dataset} for details of the handshape dataset and its construction.

For the handshape recognition model, since we operate directly on skeletal data, we adopt a straightforward 3-layer GCN architecture. This design is intuitive and intentionally kept shallow to mitigate overfitting, while still being expressive enough to capture the discriminative skeletal patterns of different handshapes, as shown in Figure~\ref{fig:stgcn}. We use standard cross-entropy loss for training.

It is worth noting that our model is not intended to assign a definitive handshape label to every frame in a video (nor would it be realistic to attempt to do so). This is because almost all frames contain hand configurations that  deviate in some way from the 87 canonical handshapes. These canonical handshapes are idealized abstractions, whereas in practice, handshapes often fall somewhere in between. Instead, the purpose of our classifier is to extract handshape features by estimating the degree of similarity between the observed handshape and handshapes in the canonical set.
\vspace{-2pt}
\subsection{Multimodal Module}
\vspace{-2pt}
The goal of the multimodal module is to incorporate the additional per-frame information from by the handshape classification module into the segmentation network. To do this, we introduce a unified cross-attention-based fusion module into the segmentation network; see Figure~\ref{fig:stgcn}.
\vspace{-12pt}
\subsubsection{Self-Attention}  
\vspace{-5pt}
For context, we begin by reviewing the self-attention mechanism\cite{vaswani2017attention}. Given an input sequence \(X \in \mathbb{R}^{T \times d}\), the attention is computed by projecting it into query, key, and value spaces via learned matrices \(W_Q, W_K, W_V\), and computing:

\begin{equation}
\text{Attn}(Q, K, V) = \text{softmax}\left( \frac{QK^\top}{\sqrt{d}} \right) V,
\end{equation}
\noindent
where \(Q = W_Q X\), \(K = W_K X\), and \(V = W_V X\), \(\sqrt{d}\) is the scaling factor.
\vspace{-15pt}
\subsubsection{Cross-Attention Module}  
We treat the segmentation features \(X^{\text{seg}} \in \mathbb{R}^{T \times d}\) as the source of the query, and concatenate the keys and values from both the segmentation and handshape streams to form the joint attention pool.

Let \(X^{\text{hand}} \in \mathbb{R}^{T \times d}\) be the pooled handshape features. We compute:
\vspace{-5pt}
\[
Q_c = W_Q X^{\text{seg}},\quad
K_c = \begin{bmatrix}
W_K X^{\text{seg}} \\
W_K X^{\text{hand}}
\end{bmatrix},\quad
V_c = \begin{bmatrix}
W_V X^{\text{seg}} \\
 W_V X^{\text{hand}}
\end{bmatrix}
\]
\noindent
The resulting cross-attention is then used to update the segmentation representation as follows:
\begin{equation}
\hat{X}^{\text{seg}} = X^{\text{seg}} + \sigma(g) \cdot \text{softmax}\left( \frac{Q_cK_c^\top}{\sqrt{d_c}} \right) V_c.
\end{equation}

Here, \(\sigma(g)\) is a learnable gating scalar that modulates the fused output, and \(\sqrt{d_c}\) is the scaling factor used in dot-product attention~\cite{vaswani2017attention}.

This formulation enables the model to leverage both internal temporal dynamics and relevant external cues from the handshape stream, while the gating mechanism ensures that cross-modal information is selectively integrated in a learnable and data-dependent way.

\subsection{Sign Recognition Using Boundaries from Segmentation}

The predicted boundaries are used to segment the video into sign clips, among which only those segments that have been successfully associated with a ground truth sign and for which we have at least 6 examples in our training dataset are passed to a sign recognition model. Specifically, we integrate into our framework a state-of-the-art sign classification model \cite{zhou2024multimodal}, which is used to predict the gloss label for each segment correctly identified by our boundary detection model. Given the output of the boundary detection model, we first identify the corresponding segments from the input sequence. These segments are then passed to the sign language classification model for recognition, laying the groundwork for a future continuous sign language recognition pipeline \cite{zhou2024multimodal}. We use this to show how good boundary segmentation is for sign recognition.

\section{Experiments}

\subsection{Dataset Construction}

\subsubsection{Handshape Recognition Dataset}
\label{sec:handshape_dataset}

We integrated 3 different ASLLRP \cite{neidle2022asl} ASL datasets---ASLLVD\cite{asllvd-data,neidle2012challenges}, DSP\cite{dsp_signs2024}, ASLLRP Sentences \cite{asllrp_sentences2025} (henceforth ASLLRP-S)---to construct a unified dataset for handshape recognition. We also used a set of videos collected specifically to illustrate the 87 handshapes in motion from different camera angles: NCSLGR \cite{dataset:databases2007volumes}. For each handshape category, the NCSLGR videos provide synchronized multi-view video recordings for the 87 linguistically defined handshapes. We sampled at least 20 frames per handshape from different camera angles, enabling the model to learn cross-view representations of each handshape. In addition, we augmented the dataset by sampling 30 instances per handshape from the 3 other sign language datasets. For rare handshapes that do not meet the 30-sample threshold, we collected as many representative samples as possible. Through this process, we constructed a comprehensive handshape dataset that captures rich visual diversity across categories and views.
\vspace{-16pt}
\subsubsection{Sign Recognition Dataset}
\vspace{-2pt}
For continuous sign data, we adopt the current state-of-the-art sign recognition method \cite{zhou2024multimodal}. We constructed a unified sign classification dataset by integrating video clips from 6 different sources. We use 4 collections with citation-form ASL signs—WLASL \cite{li2020word}, RIT \cite{rit2024}, ASLLVD \cite{asllvd-data,neidle2012challenges}, and DSP \cite{dsp_signs2024}—as well as 2 datasets in which signs had been extracted from ASL sentences based on manual annotations carried out using SignStream® (see http://www.bu.edu/asllrp/SignStream/3)---ASLLRP-S \cite{asllrp_sentences2025} and DSP Sentences \cite{dsp_sentences2025} (hereafter DSP\_S). 
 
As previously mentioned, it is critically important that the datasets for this research all make use of consistent glossing conventions.  To ensure this, we have modified some of the gloss ID labels for the WLASL dataset to conform to the conventions used for the other datasets \cite{neidle2022resources, dafnis2022isolated, neidle2022alternative, neidle2022revised}. Note also that we are relying on “class label” ID glosses to represent signs, which group together minor variants in sign production \cite{neidle2022documentation, neidle2022revised}.  

We exclude  from the sentence datasets signs designated as "hidden".  This label identified false starts and  signs that otherwise seriously deviate from their canonical articulation.  For further details, see https://dai.cs.rutgers.edu/asllvd/signbank/Additional-information-6-29-25.pdf. We used the hidden signs for purposes of segmentation, but excluded them from the sign recognition research.   We further restrict the recognition research to signs that were correctly segmented and for which our training set includes at least 6 examples.

\vspace{-10pt}
\subsubsection{Boundary Detection Dataset}

For boundary detection, we use all of the pre-segmented clips from the ASLLRP-S \cite{asllrp_sentences2025} dataset, including the “hidden signs,” as described in the previous section.  The segmentation was done based on the annotations for start and end frames of each sign. 


\subsection{Experiment Setup}

\begin{figure}[t]
\centering
\includegraphics[width=1\linewidth]{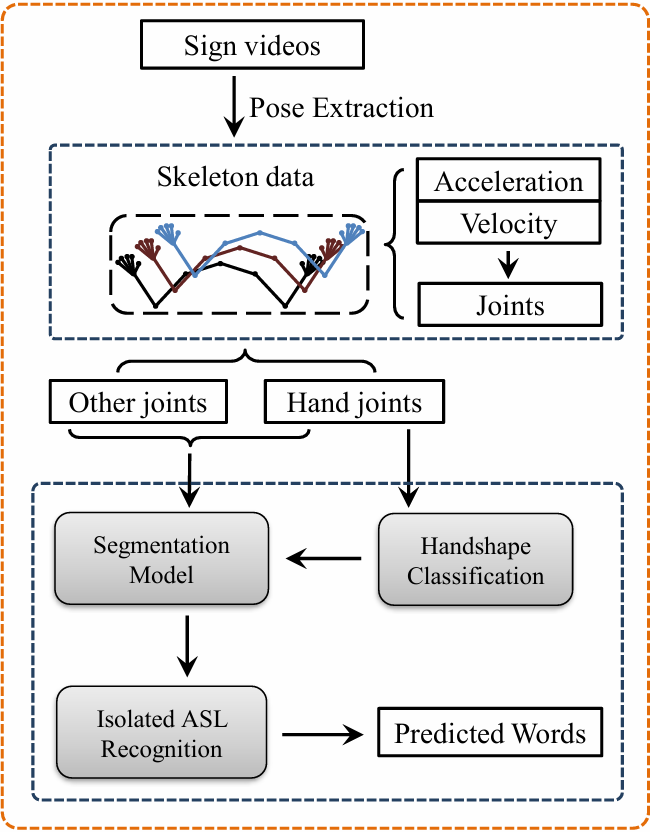}
\caption{Overview of the experimental procedure. We begin by extracting 2D skeletal keypoints from sign language videos using a pose estimation tool. We use velocity and acceleration information to augment the joint feature for each frame. These are then separated into two branches: hand joints are fed into a pretrained handshape classification model, while all joints are used as input to a segmentation model. The outputs of both branches are fused to produce improved segmentation boundaries. For sign recognition, the segmented sign clips are further passed into a sign recognition model trained on both citation-form signs and signs segmented from continuous signing to produce final sign predictions.}
\label{fig:pipeline}
\end{figure}

\subsubsection{Data Preprocessing}

For our constructed datasets, we first extract skeletal features, including keypoints of the torso, arms, and hands, from RGB frames using the AlphaPose model\cite{fang2022alphapose}. We also incorporate velocity and acceleration information derived from the keypoint sequences to enrich the temporal dynamics. For both the boundary detection and continuous sign language recognition tasks, we employ a skeletal graph consisting of 27 nodes. In contrast, for the pretrained handshape recognition model, we utilize a skeletal graph constructed solely from hand joints. 

In addition, we apply normalization and data augmentation techniques inspired by \cite{jiang2024signclip, bohavcek2022sign, zhou2024multimodal}. Specifically, we normalize the skeleton keypoints to the range \([-1, 1]\) to enhance cross-dataset transferability. To further improve model robustness, we also apply random horizontal flipping as a data augmentation strategy during training.
\vspace{-11pt}
\subsubsection{Experiment Details}

To ensure a fair comparison with the baseline \cite{renz2021sign}, we adopt the same 4:1 train-test split ratio and randomly partition the ASLLRP-S dataset. As shown in Figure \ref{fig:pipeline}, our overall pipeline consists of skeletal feature extraction, a pretrained handshape classification model, and segmentation and recognition modules. For the boundary detection task, the model is trained only up to the segmentation module to predict temporal boundaries. For the sign language recognition task, the recognition results of each correctly predicted segment are used to further optimize the boundary detection module. All models are trained on 8 NVIDIA RTX 8000 GPUs.
\vspace{-10pt}
\subsubsection{Evaluation Metrics}

We define different evaluation metrics for different tasks. For boundary detection, to facilitate comparison with prior related research, we adopt mF1B (mean F1 score of boundary) and mF1S (mean F1 score of segment) proposed by Renz \cite{renz2021sign}. The mF1B metric measures the mean F1 score for boundary detection, where a predicted boundary is considered correct if its distance to a ground-truth boundary is less than a predefined threshold. The mF1S metric measures the mean F1 score for sign segmentation, where a predicted sign segment is considered correct if its intersection-over-union (IoU) with a ground-truth segment exceeds a predefined threshold. We adopt the same threshold settings as the baseline paper \cite{renz2021sign}, to ensure fair comparison.

We also propose a more intuitive evaluation metric. Specifically, we introduce a tolerance-based matching criterion to evaluate the alignment between predicted and ground-truth sign segments, which we will refer to as Boundary Tolerance. A predicted segment is considered correct iff both its start frame and end frame are within a predefined tolerance of the ground-truth start and end frames, respectively. The tolerance is adaptive to the duration of the sign: ±2 frames for signs with duration of 1-5 frames, ±3 frames for signs with 6–10 frames, ±4 frames for signs with  11–23 frames, and ±5 frames for longer signs. This adaptive metric ensures that both boundaries are accurately captured within a reasonable margin. Allowing for some divergence takes account of the fact that neighboring frames may be very similar in appearance; there may also be a little variability in manual annotations for the same reason.  Thus, this tolerance allows for reasonable evaluation.

For the CSLR task, we do not employ metrics such as WER (Word Error Rate), which are commonly applied in this field \cite{guo2024gloss}. Instead, we use our tolerance-based matching criterion, Boundary Tolerance, to determine whether we have achieved a successful segmentation.  We consider a predicted segment to be aligned with a sign in the ground truth if both the start and end frames of the two signs are within the tolerance range.

We then compute the top-1 recognition accuracy across all predicted sign segments that were successfully matched based on this criterion. This proposed evaluation metric is motivated by 2 key considerations: (1) since our pipeline first detects boundaries and then performs recognition on individual segments, the issue of sign order and substitution, as penalized in WER, is not applicable in our setting; (2) this metric allows us to evaluate recognition accuracy in conjunction with different boundary detection settings.

\vspace{-2pt}
\subsection{Evaluation Benchmark}
\vspace{-2pt}
We separately evaluate our model on 3 tasks: boundary detection, sign recognition from continuous signing, and handshape classification.
\vspace{-3pt}
\subsubsection{Segmentation Comparison}

We compare our segmentation results with previous work \cite{renz2021sign}, although our experiments are carried out on the ASLLRP-S dataset, which is more complex and challenging than the PHOENIX14 dataset \cite{forster2014extensions}. Specifically, ASLLRP-S contains approximately 2,000 distinct sign classes and fewer than 20,000 annotated sign clips, whereas PHOENIX14 includes around 1,200 sign classes with approximately 23,000 annotated sign clips.

Despite the fact that our dataset is significantly more complex, we adopt the same training-testing split ratio used in prior work and directly compare segmentation performance. In Table \ref{tab:segmentation_comparison}, \cite{schembri2013building} reports a segmentation performance of 61.12 (mF1B) on the BSL-1K test dataset. For the PHOENIX14 dataset, \cite{renz2021sign} achieves an mF1B of 71.50.

In contrast, our model achieves substantially stronger segmentation performance on the more challenging ASLLRP-S dataset, with an mF1B of 79.40.  Given that ASLLRP-S presents greater complexity than PHOENIX14 in terms of vocabulary size and class distribution, these results demonstrate the robustness and effectiveness of our proposed segmentation approach.





\begin{table}[h]
\centering
\begin{tabular}{llcc}
\toprule
\textbf{Method} & \textbf{Dataset} & \textbf{mF1B} & \textbf{mF1S} \\
\midrule
 \textsc{$RF_{GEO}$\cite{farag2019learning}} & \textsc{BslCorpus\cite{schembri2013building}}     & 68.68 & 41.71 \\
 \textsc{$I3D,TCN$\cite{renz2021sign}} & \textsc{Phoenix14}     & 71.50 & 52.78 \\
 \textsc{$Our~MHB$} & \textsc{ASLLRP-S} & \textbf{79.40} & \textbf{58.26} \\
\bottomrule
\end{tabular}
\caption{State of the art comparison. We compare to the geometric features proposed by \cite{farag2019learning} on BSLCORPUS and the I3D + MS-TCN method proposed by \cite{renz2021sign} on Phoenix14. We show that our method significantly outperforms others on the different metrics.}
\label{tab:segmentation_comparison}
\end{table}
\vspace{-9pt}
\subsubsection{Sign Recognition}

In the sign recognition task, we adopt a different setting than was used for the standalone boundary detection task, as described earlier. To the best of our knowledge, this is the first work to evaluate sign recognition on continuous sign language datasets using segments derived from predicted boundaries. Therefore, we adopt the following evaluation metric introduced in our work. During evaluation, We assess performance from two perspectives: boundary detection and overall recognition.

For boundary detection, we adopt our Boundary Tolerance matching criterion. Specifically, we report 2 metrics to evaluate segmentation performance: (i) the proportion of ground-truth signs that have predicted segments that match under our Boundary Tolerance criterion; (ii) the proportion of predicted segments that match a ground-truth sign under our Boundary Tolerance criterion. This provides a clear assessment of segmentation accuracy, as shown in Table \ref{tab:boundary_outcomes}.

\begin{table}[h]
\centering
\begin{tabular}{lccc}
\toprule
\textbf{Segment type} & \textbf{Matched} & \textbf{Total} & \textbf{Proportion} \\
\midrule
Ground-truth signs & 3783 & 6595 & 57.4\% \\
Predicted segments & 3783 & 6477 & 58.5\% \\
\bottomrule
\end{tabular}
\caption{Segmentation accuracy evaluated based on our tolerance-based matching criterion. For each segment type, \textbf{Matched} denotes the number of sign segments that have corresponding sign segments under our matching criterion, \textbf{Total} is the total number of segments in the ground-truth or predicted sign segments, and \textbf{Proportion} is the ratio of matched to total.}
\label{tab:boundary_outcomes}
\end{table}

For CSR, we evaluate the reliability of our predicted segmentation boundaries by conducting experiments under varying thresholds k for the minimum number of training samples per sign class. Evaluation is restricted to predicted sign segments that can be aligned to ground-truth sign segments under our Boundary Tolerance criterion. For varying values of k, we report the top-1 sign recognition accuracy in 
Table \ref{tab:recognition_eval2}, which shows that top-1 recognition accuracy on boundary-matched segments increases consistently as the minimum sample threshold k rises, ranging from 80.2\% to 83.3\%. These results demonstrate that our predicted segmentation boundaries provide reliable alignment with ground truth, and that recognition performance is  strong and improves with stricter data thresholds, confirming the robustness of our approach for downstream sign recognition tasks.




\begin{table}[h]
\centering
\begin{tabular}{lcc}
\toprule
\textbf{Threshold $k$} & \textbf{Top-1 Accuracy} \\
\midrule
~~~~~~~~~6   & 80.23\% \\
~~~~~~~~10  & 80.86\% \\
~~~~~~~~15  & 81.67\% \\
~~~~~~~~20  & 82.17\% \\
~~~~~~~~30  & \textbf{83.30\%} \\
\bottomrule
\end{tabular}
\caption{Sign recognition results on the ASLLRP-S dataset under different thresholds $k$, where $k$ denotes the minimum number of training samples required per class. Top-1 Accuracy is the recognition accuracy based on the Top-1 prediction for each segment.}
\label{tab:recognition_eval2}
\vspace{-6pt}
\end{table}

\subsection{Ablation Study}

We also conduct an ablation study; see Table~\ref{tab:segmentation_ablation}. When both the handshape classification module and the multimodal fusion module are removed, leaving only the spatio-temporal convolutional segmentation model, we observe a small but significant performance drop. The model still outperforms existing state-of-the-art methods, demonstrating the strength of our base segmentation framework.

\begin{table}[h]
\centering
\begin{tabular}{llcc}
\toprule
\textbf{Method} & \textbf{mF1B} & \textbf{mF1S} \\
\midrule
 \textsc{$I3D,TCN$\cite{renz2021sign}} & 71.50 & 52.78 \\
 \textsc{$Ours~without~handshape~module$}  & 77.29 & 56.98 \\
 \textsc{$Ours~with~handshape~module$} & \textbf{79.40} & \textbf{58.26} \\
\bottomrule
\end{tabular}
\caption{Ablation study for the boundary detection task. We compare to the I3D + MS-TCN method proposed by~\cite{renz2021sign} on the Phoenix14 dataset. The 2nd row shows our method without the handshape module, and the 3rd  presents the full model.  Best performance is achieved when the handshape module is incorporated.}
\vspace{-3pt}
\label{tab:segmentation_ablation}
\end{table}

\subsection{Qualitative Analysis}
\begin{figure}[!h]
	\includegraphics[width=0.475\textwidth]{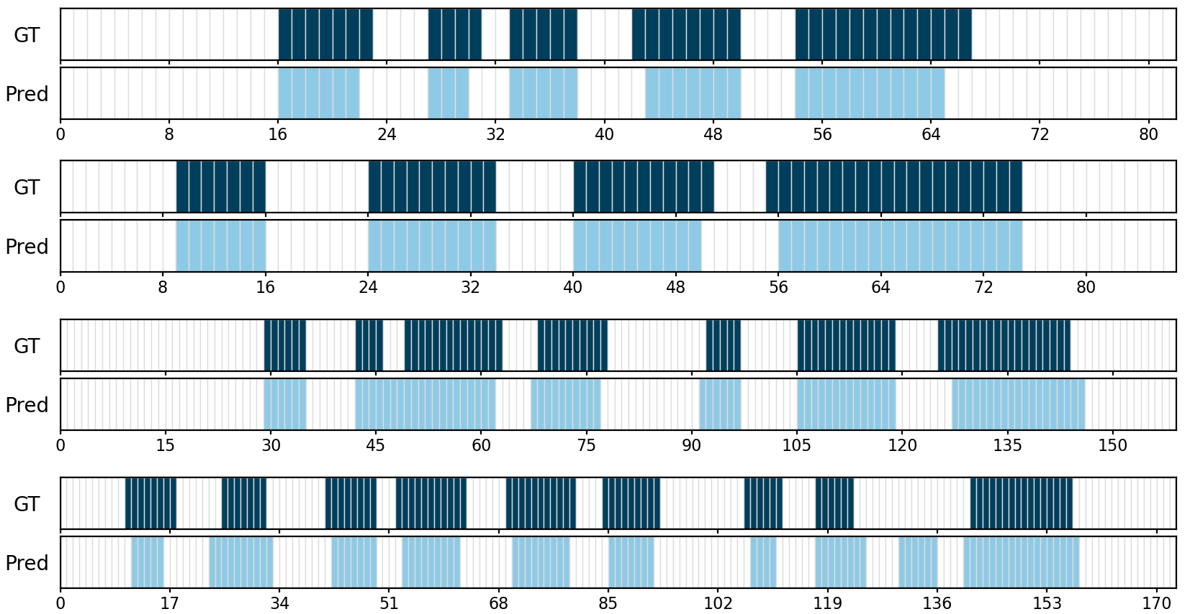}
    \captionsetup{skip=1pt}
    \vspace{1pt}
	\caption{Qualitative results on the ASLLRP-S dataset. Each pair of rows corresponds to one {ASL} sentence video: the top row in each pair shows the ground-truth segmentation (GT), and the bottom row shows the model prediction (Pred). Blue bars indicate sign segments, and white indicates background or non-sign regions. It can be seen that predicted segments from the model closely align with the ground truth, especially in the case of shorter sentences (as shown in the first and second pairs). In longer sentences, although some segments may be missed (as shown in the third pair) or two adjacent segments may occasionally be merged into one (as shown in the fourth pair), the model  delivers reasonably accurate segmentation performance overall. This highlights the model’s ability to detect temporal boundaries across sentences with different numbers of signs.}
	\label{fig:qualitative}
\end{figure}

Figure \ref{fig:qualitative} presents qualitative results on the ASLLRP-S dataset, covering both long and short sign sentences as well as corresponding ablation studies. It can be observed that our method achieves accurate segmentation in both cases, particularly excelling on short sequences. For long sentences, although some boundary errors remain, the overall segmentation performance remains precise and consistent.

\section{Conclusion}\label{sec:conclusion}

Here we address the task of boundary detection for CSLR and propose a novel pipeline for continuous sign language recognition. We integrate a pre-trained handshape classification model into a pre-trained spatio-temporal convolution segmentation module, enhancing the model’s ability to detect sign boundaries. Furthermore, we design a unified recognition framework by combining an existing state-of-the-art isolated sign language recognition model with our boundary detection model.

We construct a merged ASL handshape classification dataset by combining several existing ASL datasets. We evaluate our method on the ASLLRP-S dataset, measuring  boundary detection performance and the accuracy of recognition of successfully segmented signs.  The results demonstrate that our approach significantly outperforms existing baselines in boundary detection. Ablation studies and other experiments confirm that accurate boundary modeling contributes to the overall recognition performance.


\vspace{3pt}
\noindent\textbf {Limitations and Future Work}

In the current research, we are only paying attention to 1-handed signs produced with the dominant hand and to 2-handed signs; we are NOT paying any attention to one-handed signs produced on the non-dominant hand.  We have also focused on some specific parameters that are relevant to detection of sign boundaries.  Extending the focus to include additional relevant parameters will be for future research.

Our long-term goal is an end-to-end system for CSL segmentation and recognition. Future research will focus on distinguishing types of signs in continuous signing \cite{yanovich2016detection}, and targeting specific recognition strategies as appropriate.  In addition, future research will aim to take advantage of linguistic constraints applicable to different sign types, e.g., the dependencies between start and end handshapes and handshapes on the two hands in lexical signs \cite{dilsizian2014new, thangali2011exploiting}.




{\small
\bibliographystyle{ieee_fullname}
\bibliography{egbib}
}

\end{document}